\documentclass[runningheads]{llncs}

\usepackage[accsupp]{axessibility}  

\usepackage{wrapfig}
\usepackage{comment}
\usepackage[T1]{fontenc}
\usepackage{multirow}

\usepackage{comment}

\usepackage{graphicx}
\usepackage{wrapfig}
\usepackage{float}

\usepackage{booktabs}                   
\usepackage{multirow}
\usepackage{array}
\usepackage{paralist}
\usepackage{enumitem}

\usepackage{graphicx}                  
\usepackage{mathtools}
\usepackage{amssymb,amsmath}   

\usepackage[pagebackref=true,breaklinks=True,colorlinks=true,bookmarks=false,citecolor=blue,linkcolor=blue]{hyperref}

\usepackage{pifont}
\usepackage{xspace}
\usepackage{diagbox}
\usepackage{xcolor}
\usepackage[export]{adjustbox}
\usepackage{sidecap}

\def\eg{e.g.,~}               
\def\ie{i.e.,~}               

\newcommand{\secref}[1]{Section~\ref{sec:#1}}
\newcommand{\figref}[1]{Figure~\ref{fig:#1}} 
\newcommand{\tabref}[1]{Table~\ref{tab:#1}}
\newcommand{\eqnref}[1]{Eq.~\ref{eq:#1}}

\long\def\ignorethis#1{}

\newlength\paramargin
\newlength\figmargin
\newlength\subfigmargin
\newlength\secmargin
\newlength\subsecmargin
\newlength\tabmargin
\newlength\eqmargin

\definecolor{bittersweet}{rgb}{1.0, 0.44, 0.37}

\newcommand{\EC}{\textsc{EclipSE\text{ }}}

\begin{document}
\pagestyle{headings}
\mainmatter
\def\ECCVSubNumber{735}  

\title{\textsc{EclipSE}: Efficient Long-range Video\\ Retrieval using Sight and Sound} 

\titlerunning{ECLIPSE}
%
\author{
Yan-Bo Lin \and
Jie Lei \and
Mohit Bansal  \and
Gedas Bertasius
}
\authorrunning{Lin et al.}
%
%
\institute{Department of Computer Science \\
University of North Carolina at Chapel Hill\\
\email{\{yblin,jielei,mbansal,gedas\}@cs.unc.edu}
}

\maketitle

\begin{abstract}
We introduce an audiovisual method for long-range text-to-video retrieval. Unlike previous approaches designed for short video retrieval (e.g., 5-15 seconds in duration), our approach aims to retrieve minute-long videos that capture complex human actions. One challenge of standard video-only approaches is the large computational cost associated with processing hundreds of densely extracted frames from such long videos. To address this issue, we propose to replace parts of the video with compact audio cues that succinctly summarize dynamic audio events and are cheap to process. Our method, named \EC (Efficient CLIP with Sound Encoding), adapts the popular CLIP model to an audiovisual video setting, by adding a unified audiovisual transformer block that captures complementary cues from the video and audio streams. In addition to being $2.92\times$ faster and $2.34\times$ memory-efficient than long-range video-only approaches, our method also achieves better text-to-video retrieval accuracy on several diverse long-range video datasets such as ActivityNet, QVHighlights, YouCook2, DiDeMo, and Charades. Our code is available at \url{https://github.com/GenjiB/ECLIPSE}
\end{abstract}

\section{Introduction}
\vspace{\secmargin}

Fueled by the growing availability of video data, the last few years have witnessed remarkable progress in text-to-video retrieval~\cite{iccv21_TeachText,iccv21_HiT,iccv21_Frozen,cvpr21_T2VLAD,cvpr21_SemSim,eccv20_MMT,wacv22_MaskModality,emnlp21_videoclip}. However, modern video retrieval systems are predominantly designed for very short videos (\eg 5-15 seconds in length). In contrast, the majority of real-world videos often capture complex human actions, which may last several minutes or even hours. For example, consider yourself performing a complex activity of making Japanese Souffle Pancakes, which may take a couple of hours. In a scenario when you forget some of the steps in the recipe, it would be helpful to retrieve a relevant several-minute-long video segment demonstrating how to perform those steps. However, the traditional short-range video retrieval models would struggle with this task due to their inability to analyze longer videos. Combining the strengths of audio and video modalities, we aim to address this limitation by proposing an efficient audiovisual text-to-video retrieval system focused on long-range videos. 
\begin{figure}[t!]
    \centering
    \includegraphics[width=0.99 \linewidth]{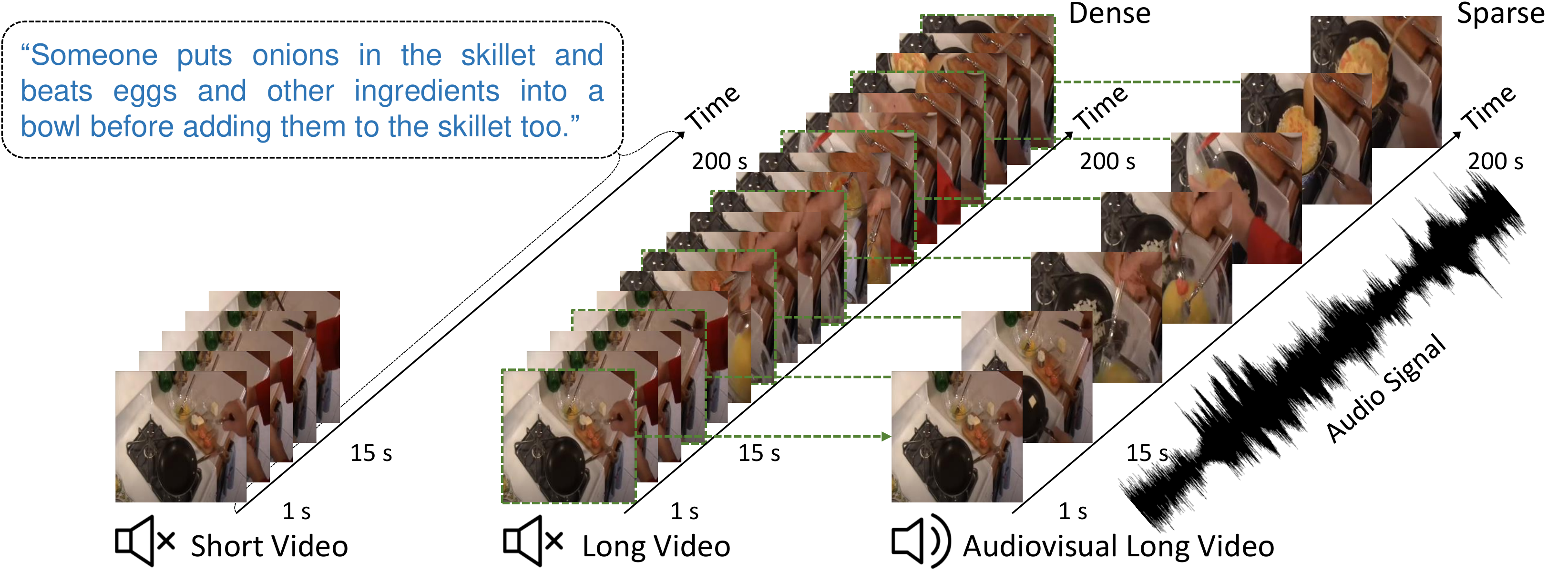}
	\vspace{\figmargin}
    \caption{Comparison of different high-level frameworks for long-range text-to-video retrieval. Most traditional text-to-video retrieval methods (\textbf{Leftmost Column}) are designed for short videos (e.g., 5-15 seconds in duration). Adapting these approaches to several-minute long videos by stacking more input frames (\textbf{Middle Column}) is impractical due to excessive computational cost. Instead, our proposed framework operates on sparsely sampled video frames and dense audio cues, which are cheaper to process  (\textbf{Rightmost Column}). In addition to being more efficient, our framework also achieves higher text-to-video retrieval accuracy than standard video-only approaches. 
    }
    \label{fig:teaser}
    \vspace{\figmargin}
\end{figure}

\begin{wrapfigure}{r}{0.45\linewidth}
    \centering
    \includegraphics[width=1\linewidth]{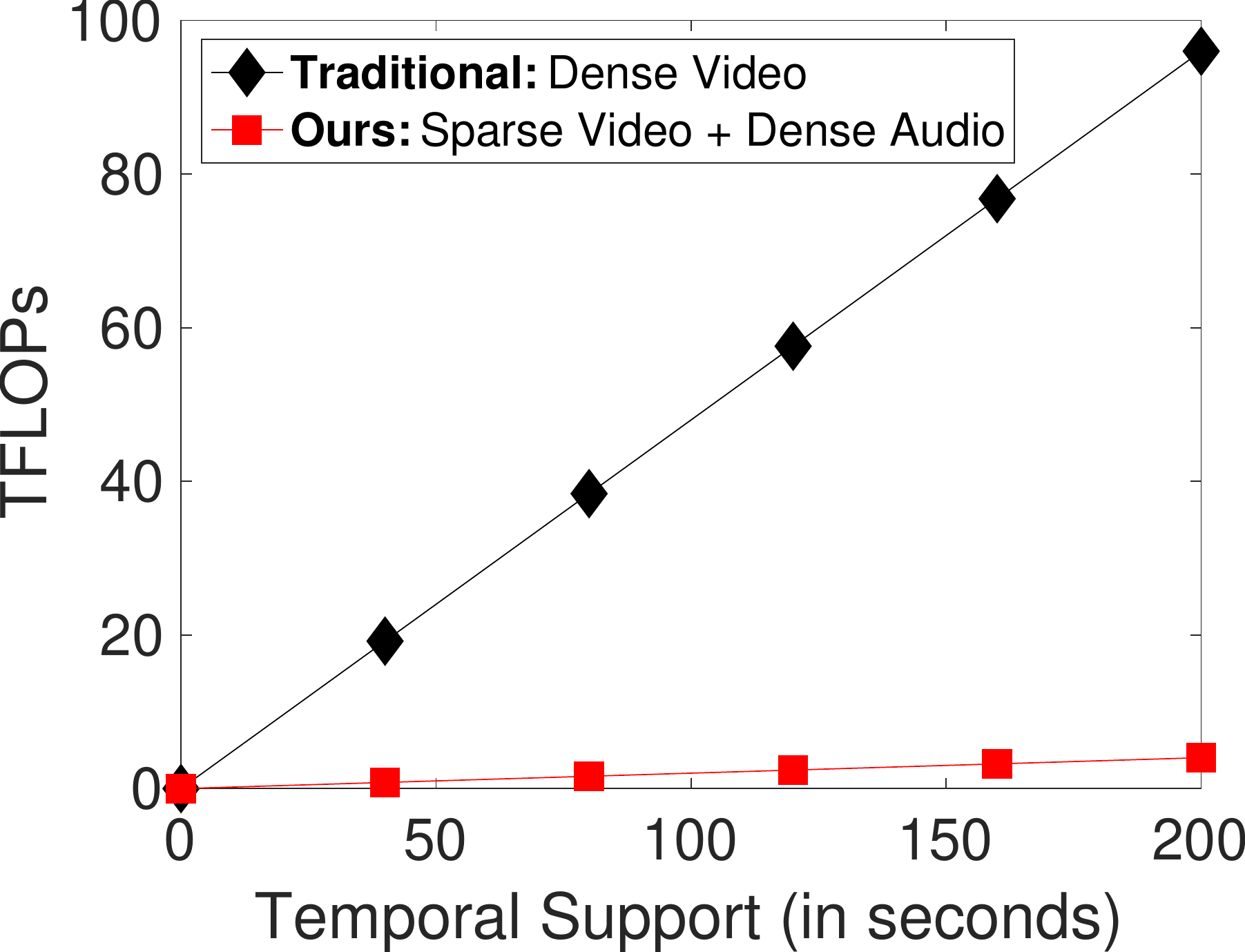}
    \caption{Our audiovisual framework scales to long videos more efficiently than dense video-only approaches. }
    \label{fig:teaser_support}
\end{wrapfigure}

Among prior vision-and-language methods~\cite{iccv21_HiT,iccv21_TeachText,cvpr21_clipbert,old_arxiv16_Word2visualvec,aaai15_joint,arxiv14_unify,arxiv_clip2video,arxiv_clip2tv}, CLIP~\cite{icml21_clip} stands out as one of the most widely adopted models. Several recent approaches extended CLIP to video~\cite{arxiv_clip4clip} by independently processing individual video frames and then averaging their predictions across time. However, these approaches are often impractical in the long-range video setting because of the large computational cost required to process hundreds of densely extracted video frames (See~\figref{teaser_support}). Furthermore, while video modality is rich in the information it stores, it also has high informational redundancy (i.e., the video content often changes little in neighboring frames). In contrast, audio can compactly capture information related to human actions~\cite{nips21_bottleneck,kazakos2021MTCN}, objects~\cite{av_eccv18_obj_that_sound,cvpr22_ready_av_pretrained,cvpr22_ready_audio_object},  scenes~\cite{av_nips16_soundnet,av_nips20_xdc} and other complex events~\cite{my_nips21} while also being cheaper to process~\cite{cvpr20_listen2look} than the raw video. For instance, consider a video of a person frying the eggs in a pan. In this example, most of the relevant visual information (e.g., kitchen stove, pan, eggs, etc.) can be captured in just a few video frames, while the temporal dynamics in the scene can be succinctly encoded in the audio stream (e.g., the sounds of the eggs sizzling in a pan, etc.).

Based on this motivation, we introduce \textsc{EclipSE}, an \textsc{E}fficient CLIP with \textsc{S}ound \textsc{E}ncoding. Instead of processing many densely-extracted frames from a long video (the middle column in Figure~\ref{fig:teaser}), our framework leverages complementary audio and video cues by operating on sparsely sampled video frames accompanied by dense audio (the rightmost column in Figure~\ref{fig:teaser}). We demonstrate that compared to dense video-only approaches, our framework is not only more efficient but it is also more accurate. 

Our approach adapts CLIP to long-range videos by incorporating a dual-pathway audiovisual attention block into every layer of the Transformer backbone. Such a cross-modal attention mechanism allows our model to (i) incorporate long-range temporal cues from the audio stream into the visual representation, and (ii) conversely inject rich visual features from the video modality into audio representation for improved audio feature expressivity. Such bi-directional exchange of information ensures that both modalities benefit from each other in order to maximize the performance of the downstream application (i.e., long-range text-to-video retrieval). Additionally, we demonstrate that our audiovisual attention block can be easily incorporated into pretrained transformer models such as CLIP~\cite{icml21_clip} without re-training the new model from scratch.

We validate \EC on several diverse long-range video retrieval benchmarks and show that it achieves state-of-the-art results on ActivityNet~\cite{iccv17_activitynet}, QVHighlights~\cite{nips21_qvhighlights}, DiDeMo~\cite{didemo}, YouCook2~\cite{youcook2}, and Charades~\cite{eccv16_charades} while being $2.92\times$ faster and $2.34\times$ memory-efficient than long-range video-only methods.

In summary, our contributions are threefold. First, we propose \textsc{EclipSE}, an audiovisual adaptation of CLIP that leverages complementary video and audio cues for long-range video retrieval. Second, we demonstrate that compared to long-range video-only approaches, our audiovisual framework leads to better video retrieval results at a reduced computational cost. Lastly, we provide comprehensive ablation studies investigating the success factors of \EC. 

\section{Related Work}
\vspace{\secmargin}

\textbf{Text-to-Video Retrieval.}
The association of text descriptions and videos provides rich supervisory signals for developing robust text-to-video retrieval systems.
Self-supervised learning approaches in this area achieve impressive results using contrastive loss \cite{iccv21_Frozen,iclr21_supportset,aaai21_noise,cvpr20_uncurated,iccv19_mmen,cvpr22_ready_bridge,cvpr22_ready_COTS}, masked language modeling \cite{cvpr22_ready_object_aware,cvpr20_actbert,emnlp20_hero,eccv18_joint,cvpr17_video_captioning_retrieval}, or masked feature prediction~\cite{axriv20_univl}.
Additionally, several prior methods propose to incorporate rich audio/speech information for video-and-text representations learning, either by fusing cross-modal signals~\cite{eccv20_MMT,bmvc19_ce,icmr18_multimodal_cues} or masking inputs from different modalities during training~\cite{wacv22_MaskModality,arxiv18_mmoe}.
Furthermore, with large-scale pre-training on millions of image and text pairs, CLIP~\cite{icml21_clip} has achieved impressive results on a wide array of vision-and-language tasks.
Recently, CLIP-based approaches \cite{arxiv_CAMoE,arxiv_multi_quary,arxiv_clip4clip,cvprw21_MDMMT,mcpr21_clip,arxiv_clip2video,arxiv_clip2tv,cvpr22_ready_xpool,arxiv22_ready_clip} have also been used in video by aggregating image-level outputs across different time steps.

Unlike these prior methods, which are designed for short-range videos (e.g., 5-15s), we aim to design an audivosual framework for retrieving long videos (e.g., several minutes in length). Compared to the existing CLIP-based approaches, which are difficult to adapt to long videos due to the large computational cost of processing many densely-extracted video frames, we propose to leverage compact audio cues in order to reduce the need for the costly video modality.
This enables efficient adaptation of CLIP to long video retrieval. 

\textbf{Audiovisual Learning.} Audio and video synchronization is commonly used for self-supervised audio-visual  learning~\cite{av_eccv18_Owens,av_nips18_coop,av_iccv17_look,av_eccv18_obj_that_sound,av_nips16_soundnet,av_eccv16_abSound,av_nips20_xdc,av_nips20_CrossLabelling,av_iclr21_activeContrastive,av_cvpr21_agreementAVID,av_cvpr21_RAVID}. Aside from self-supervised learning, many recent methods were proposed for audio-visual event classification~\cite{my_icassp,cvpr20_listen2look,nips21_bottleneck,my_nips21,nips21_av_globallocal}. Furthermore, the recent popularity of Transformers~\cite{cvpr22_ready_audio_adaptive,iclr21_vit,my_accv20_av-trans,nips17_attention} have enabled a wide array of architectures for jointly modeling audio and video data~\cite{arxiv_everyatonce,arxiv_merlot_reserve,nips21_vatt,arxiv_connectAT,avt_nips20_VersatileNet,iccv21_multimodal_cluster,my_aaai21_avconsistency}. Compared to these prior approaches, our approach focuses on efficient long-range text-to-video retrieval. Specifically, we aim to leverage audio cues in order to reduce the computational cost of processing long videos.  

\textbf{Long Sequence Modeling.} Recent work in the natural language processing (NLP) domain~\cite{wang2020linformer,choromanski2020rethinking,patrick2021keeping} proposed to approximate the self-attention operator for long sequence modeling. While these approaches are effective in NLP, they are still very costly in the video domain due to the high dimensionality of video inputs. Furthermore, as demonstrated by recent work in the video domain~\cite{patrick2021keeping}, such approximation techniques lead to a substantial accuracy drop while producing limited efficiency gains for video recognition tasks. Additionally, we note that these approximation mechanisms are often incompatible with pretrained vision-and-language models such as CLIP (due to different network architectures).

\section{\textsc{EclipSE}: Efficient CLIP with Sound Encoding}

\begin{figure}[t!]
    \centering
	\includegraphics[width=1\linewidth]{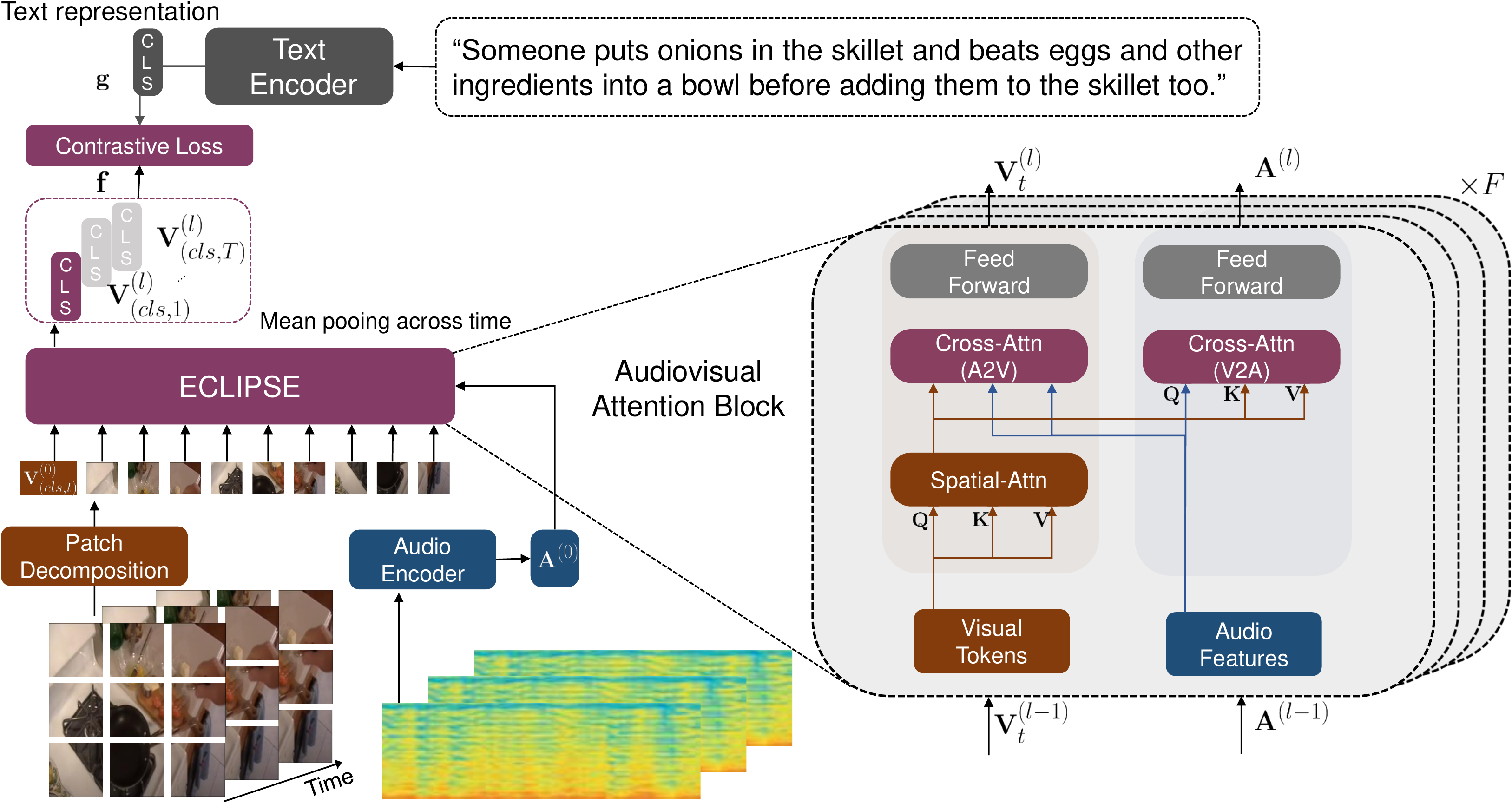}
	\vspace{\figmargin}
    \caption{
    We adapt CLIP~\cite{icml21_clip} to long-range text-to-video retrieval by adding an efficient audiovisual attention block into the Transformer architecture. First, we obtain fixed dimensional text, audio, and visual feature embeddings. Afterward, the visual and audio embeddings are fed into our \EC audiovisual backbone, which injects relevant audio information to video and vice-versa. This is accomplished using a dual-pathway audiovisual attention block (illustrated on the right), which is stacked on top of each other $F$ times. Afterward, the audiovisual video segments are aggregated using temporal pooling, and the model is optimized by maximizing the similarity between audiovisual and textual embeddings using a contrastive loss function.\vspace{-0.1cm}
    }
	\label{fig:method}
	\vspace{\figmargin}
\end{figure}

Our goal is to design an efficient framework that leverages audiovisual cues for long-range text-to-video retrieval. Instead of processing many densely-extracted frames from a long video, which is costly, our framework operates on sparsely sampled video frames accompanied by dense audio. We adapt CLIP to long-range videos by adding a dual-pathway audiovisual attention block into every layer of the Transformer backbone. Our video retrieval framework consists of three high-level components: (i) multimodal input embeddings, (ii) an audiovisual backbone for processing video and audio modalities, and  (iii) a contrastive video-to-text matching objective. Below we provide more details behind each of these components. We also illustrate our framework in Figure~\ref{fig:method}.

\subsection{Obtaining Multimodal Input Embeddings}
 
\textbf{Video, Audio and Text Inputs.}  Our framework takes audio, video, and text modalities as its inputs. For video modality, we consider video clips $X \in \mathbb{R}^{T \times H \times W \times 3}$ consisting of $T$ RGB frames of size $H \times W$, sampled uniformly from the whole input video.
For audio, we use $T$ audio spectrograms $Z \in \mathbb{R}^{T \times M \times C}$, each spanning $t$ seconds and centered around each of $T$ video frames. Here, $M$ and $C$ depict spatial spectrogram dimensions. Lastly, the text is represented as a sequence $y = (y_1, \hdots, y_L)$ where $y_i$ represents a distinct word in the textual video description and $L$ is the length of the description (i.e., the number of words).

\textbf{Video Patch Decomposition.} Following the ViT~\cite{iclr21_vit}, we decompose each frame into $N$ non-overlapping patches, each of size $P \times P$, and flatten these patches into vectors $\smash{{\bf x}_{(p,t)} \in \mathbb{R}^{3 P^2}}$ where $p=1,\hdots,N$ denotes spatial locations and $t=1,\hdots,T$ indicates a frame index.

\textbf{Video Patch Embeddings.} Video patches from each frame $\smash{{\bf x}_{(p,t)}}$ are linearly mapped into vectors $\mathbf{v}^{(0)}_{(p,t)} \in \mathbb{R}^{d}$, for $p=1 \ldots N$, and $t=1 \ldots T$. Afterward, we also augment each visual token with spatiotemporal position information as is done in~\cite{gberta_2021_ICML}. A specialized CLS token $\mathbf{v}^{(0)}_{cls}$ is prepended to the visual sequence of each frame. Finally, the embeddings $\mathbf{V}^{(0)} \in \mathbb{R}^{ T \times (N+1) \times d}$ are used as visual inputs to our \EC model.

\textbf{Audio Embeddings.} Given an audio spectrogram $Z_t  \in \mathbb{R}^{M \times C}$, an audio encoder maps it into audio embeddings $\mathbf{A}^{(0)}_t \in \mathbb{R}^{d}$ for each timestep $t=1 \ldots T$ where as before, $T$ denotes the number of video frames. We note that the audio encoder can be either a CNN~\cite{icassp17_audioset,icassp17_vggish,icassp20_vggsound} or a Transformer~\cite{interspeech21_AST,TASLP_AST}. Afterward, the audio embeddings $\mathbf{A}^{(0)} \in \mathbb{R}^{T \times d}$ are fed into \EC together with the visual tokens $\mathbf{V}^{(0)} \in \mathbb{R}^{ T \times (N+1) \times d}$.

\textbf{Text Embeddings.} We use a pretrained CLIP~\cite{icml21_clip} text encoder to embed a textual video description $y = (y_1, \hdots, y_L)$ into a textual embedding $\mathbf{g} \in \mathbb{R}^{d}$ where $\mathbf{g}$ corresponds to the CLS token of a given text sequence.

\subsection{Audiovisual Attention Block}
\label{sec:eclipse}

Although videos contain rich information, they are also redundant and costly to process. In contrast, audio is more compact and cheaper. Thus, we propose an audiovisual attention block that gradually incorporates relevant audio cues into the visual representation. Our audiovisual attention block consists of three distinct attention schemes: (i) spatial visual attention, (ii) audio-to-video attention, and (iii) video-to-audio attention. We next describe each of these attention schemes in more detail.

\textbf{Multi-Head Self-Attention.} All of our three attention schemes are implemented using a standard multi-head self-attention:

\begin{equation}
\begin{aligned}
\label{eq:att_func}
\vspace{\eqmargin}
\text{MHA}(\mathbf{Q},\mathbf{K},\mathbf{V}) =  \mathrm{Softmax}\left( \dfrac{\mathbf{Q}\mathbf{K}^\top}{\sqrt{d}}\right) \mathbf{V},
\end{aligned}
\vspace{\eqmargin}
\end{equation}
where $\mathbf{Q},\mathbf{K},\mathbf{V}$ are the query, key and value matrices obtained using learnable projection weights $\mathbf{W}^{Q},\mathbf{W}^{K},\mathbf{W}^{V} \in \mathbb{R}^{d \times d}$ respectively. With this formal description of the \text{MHA} function, we can now proceed to the definitions of the three attention schemes in our audiovisual attention block.

\textbf{Spatial Attention.} In order to preserve the pretrained network structure of CLIP, we use an identical spatial attention scheme as in CLIP. Intuitively, spatial attention enables our model to obtain discriminative frame-level representation by aggregating relevant information from the visual tokens in the individual video frames.  We can implement this scheme using our previously defined MHA function as:
\begin{equation}
\begin{aligned}
\label{eq:spatial_att}
\vspace{\eqmargin}
\mathbf{S}_t^{(\ell)} = \text{MHA}(\mathbf{V}_t^{(\ell-1)}, \mathbf{V}_t^{(\ell-1)}, \mathbf{V}_t^{(\ell-1)}) + \mathbf{V}_t^{(\ell-1)}. 
\end{aligned}
\vspace{\eqmargin}
\end{equation}
Here, $\mathbf{S}_t^{(\ell)} \in \mathbb{R}^{(N+1) \times d}$ is our newly computed spatial self-attention representation for frame $t$, and $\mathbf{V}_t^{(\ell-1)}$ is a visual patch representation for frame $t$ from the previous transformer layer $l-1$, which is used as input to the transformer layer $l$. Note that in the spatial self-attention, the multi-head self-attention is applied independently for each of $T$ video frames. As discussed above, this enables us to preserve the network structure of the original CLIP model, which is essential for good text-to-video retrieval performance. For brevity, we omit the layer normalization operation, which is applied to $\mathbf{V}_t^{(\ell)}$ before feeding it to the spatial attention block. The right part of Figure~\ref{fig:method} provides a visual illustration of where spatial attention fits within our audiovisual attention block.

\textbf{Audio-to-Video Attention (A2V).} To efficiently incorporate temporal audio cues into static video frame representation, we use an audio-to-video (A2V) attention mechanism, which is also illustrated in the right part of Figure~\ref{fig:method} (labeled as Cross-Attn A2V module). This operation can be written as:

\begin{equation}
\begin{aligned}
\label{eq:a2v_att}
\vspace{\eqmargin}
\mathbf{V}_t^{(\ell)} = \text{MHA}(\mathbf{S}_t^{(\ell-1)}, \mathbf{A}^{(\ell-1)}, \mathbf{A}^{(\ell-1)}) + \mathbf{S}_t^{(\ell-1)}. 
\end{aligned}
\vspace{\eqmargin}
\end{equation}
Here, $\mathbf{A}^{(\ell-1)} \in \mathbb{R}^{T \times d}$  depicts our previously defined audio representation at layer $l-1$, and $\mathbf{S}_t^{(\ell-1)} \in \mathbb{R}^{(N+1) \times d}$ denotes a spatial video representation at timestep $t$ computed using our previously defined spatial attention block. Intuitively, the new visual representation $\mathbf{V}_t^{(\ell)}$ is computed as a weighted summation of the audio features, which enables the model to incorporate long-range audio cues into the visual features. Furthermore, because the audio representation is compact, the operation above can be implemented efficiently.

\textbf{Video-to-Audio Attention (V2A).} Conversely, to inject rich visual information into compact audio features, we use a video-to-audio (V2A) attention mechanism (illustrated in Figure~\ref{fig:method} as Cross-Attn V2A module). We implement this attention scheme as: 

\begin{equation}
\begin{aligned}
\label{eq:v2a_att}
\vspace{\eqmargin}
\mathbf{A}_t^{(\ell)} = \text{MHA}(\mathbf{A}_t^{(\ell-1)}, \mathbf{S}_t^{(\ell-1)}, \mathbf{S}_t^{(\ell-1)}) + \mathbf{A}_t^{(\ell-1)}. 
\end{aligned}
\vspace{\eqmargin}
\end{equation}
At a high level, the operation above computes a new audio feature representation for each timestep $t$ as a weighted combination of all the visual token features at timestep $t$. This allows us to improve the richness of the audio representation.

\textbf{Final Audiovisual Representation.} Following CLIP4Clip~\cite{arxiv_clip4clip}, we stack our audiovisual attention block $F$ times ($F$ typically being set to $12$). Afterward, we perform temporal pooling over the CLS tokens across all video frames, to obtain the final audiovisual representation $\mathbf{f} \in \mathbb{R}^{d}$.

\subsection{Loss Function}
\label{sec:loss_func}
We use the same contrastive video-to-text matching loss as in~\cite{arxiv_clip4clip}. Specifically,  we compute the similarity between text and video using a normalized dot product between the two embeddings $\mathbf{f}$ and $\mathbf{g}$. We consider the matching text-video pairs in a given batch as positive samples and all the other pairs in that same batch as negative samples. To train our model, we minimize the sum of the video-to-text and text-to-video matching losses~\cite{arxiv_clip4clip}.




\subsection{Implementation Details}
\label{subsec:training_details}

Our \EC follows CLIP4Clip~\cite{arxiv_clip4clip} setting where text encoder and visual encoder are initialized with CLIP weights~\cite{icml21_clip}. Specifically, we initialize the spatial attention weights with the weights from CLIP. We also use CLIP weights to initialize both of our cross-modal attention blocks. We attach zero-initialized linear projection layers to the outputs of both cross-modal attention blocks so that the initial outputs of these blocks would be set to zero. 
Unless otherwise noted, for all of our experiments, we use a ViT-B/32 with uniformly sampled 32-frame inputs spanning the whole input video.  The visual frames are extracted at $3$ fps. 
We implement \EC using Pytorch~\cite{pytorch} and conduct the training on four NVIDIA A6000 GPUs. 
For a fair comparison with the baselines, we set the batch size to 64.
For audio encoder, we use ResNet-18~\cite{resnet} pre-trained on VGGSound~\cite{icassp20_vggsound}.
We sample 10-second audio clips in the neighborhood around the sampled video frame and process the raw audio into a spectrogram as is done in~\cite{icassp20_vggsound}.
We train our model with Adam optimizer~\cite{adam} and set the learning rate to $1e-7$ for text encoder and spatial attention in~\eqnref{spatial_att}.
%
%
The frame-level \textit{CLS} tokens are averaged to obtain the final video embedding.

Furthermore, the maximum text input is set to $64$ tokens for DiDeMo and QVHighlight, and $128$ for ActivityNet Captions and YouCook2.
%
%

\section{Experimental Setup}

\subsection{Downstream Datasets}
\label{subsec:downstream_datasets}

We evaluate \EC on five diverse long-range datasets: ActivityNet Captions~\cite{iccv17_activitynet}, QVHighlights~\cite{nips21_qvhighlights}, DiDeMo~\cite{didemo}, YouCook2~\cite{youcook2}, and Charades~\cite{eccv16_charades}.

\textbf{ActivityNet Captions}~\cite{iccv17_activitynet} consists of 20,000 YouTube human activity videos, each annotated with temporally localized sentence descriptions, with a total of 100,000 sentences.
The average video length is 180 seconds, which makes this dataset well suited for verifying our model's ability to retrieve long-range videos. 
We follow~\cite{arxiv_clip4clip,eccv18_cm_hierarchical,eccv20_MMT,iccv21_HiT} to evaluate paragraph-to-video retrieval, where we concatenate all the sentence descriptions to form a paragraph.
Since there is no test set provided, we evaluate the video retrieval results on the \textit{val1} split.

\textbf{QVHighlights}~\cite{nips21_qvhighlights} contains 3,164 videos (10,148 clips)  from YouTube, covering a wide range of topics, including everyday activities in lifestyle vlog videos to social and political activities in news videos. Each video is temporally annotated with multiple text queries describing distinct spans of the video.
The average video length is around 8 minutes. The original dataset is created for moment localization and highlight detection. Here we re-purpose it for text-to-video retrieval by evaluating it in paragraph-to-video retrieval setup as ActivityNet Captions. We use the standard splits for training, validation, and testing.

\textbf{DiDeMo}~\cite{didemo} contains 10,464 Flickr videos with 40,543 temporally localized sentences. The average video length is 30 seconds. Similar to ActivityNet Caption, we evaluate paragraph-to-video retrieval on DiDeMo. We use the standard splits for training, validation, and testing.

\textbf{YouCook2}~\cite{youcook2} consists of 2,000 videos capturing 89 complex recipes with total duration of 176 hours. The average video length is 5.26 minutes. 
Each video is annotated with multiple temporally localized captions.
Similar to ActivityNet Captions, we evaluate all methods in the paragraph-to-video retrieval setting. We use standard splits for training, validation, and testing.

\textbf{Charades}~\cite{eccv16_charades} contains 9,848 videos with the corresponding textual descriptions. The average video length is about 28 seconds. We use standard train and test splits for training and testing. 


\subsection{Evaluation Metrics}

We use standard video retrieval evaluation metrics~\cite{cvpr21_clipbert,arxiv_clip4clip} such as text-to-video $R@1$, $R@5$, $R@10$, and mean rank (MnR) to validate the effectiveness of our \EC model.  Since our model is built on CLIP, which is pretrained on a large-scale image-and-text dataset~\cite{icml21_clip}, the comparisons with some of the previous methods are not directly applicable. Therefore, in all of our evaluations, we use a publicly available state-of-the-art CLIP4Clip~\cite{arxiv_clip4clip} video retrieval system as our primary baseline.

\section{Results and Analysis}

\subsection{ActivityNet Captions}

\begin{table}[t]
    \caption{
    \textbf{ActivityNet Captions.} We compare \EC with previous video retrieval methods. 
    In the column Pretrain, C,G,H,W,CW,V denote COCO Captions~\cite{COCOCaptions}, Visual Genome Captions~\cite{VisGenme}, HowTo100M~\cite{howto100m}, WIT~\cite{icml21_clip}, CC3M~\cite{cc3m}+WebVid2M~\cite{iccv21_Frozen} and VGGSound~\cite{icassp20_vggsound} datasets respectively.
    The performance is evaluated using text-to-video retrieval R@1, R@5, R@10 and MnR metrics. \EC achieves the best reported accuracy on this benchmark. We also note that using a stronger visual backbone (i.e., ViT-B/16 vs. ViT-B/32) also leads to better video retrieval performance.
    }
    \label{tab:act_stoa}
    \centering
    \setlength{\tabcolsep}{2.5pt} 
    \footnotesize
    \begin{tabular}{l cccccc}
        \toprule
        Method & Pretrain & Frames & R@1 $\uparrow$ & R@5 $\uparrow$  & R@10 $\uparrow$ & MnR $\downarrow$ \\
        \midrule
        CE~\cite{bmvc19_ce}              &  - &  -  & $18.2$ & $47.7$ &  $-$ &  $23.1$ \\
        ClipBERT~\cite{cvpr21_clipbert}  &  C+G &  40 & $21.3$ & $49.0$ &  $-$ & $-$ \\
        TT-CE~\cite{iccv21_TeachText}    &  -  &  64 & $23.4$ & $57.2$ &  $-$ &  $-$ \\
        MMT~\cite{eccv20_MMT}            &  H &  -  & $28.7$ & $61.4$ &  $-$ &  $16.0$ \\
        FiT~\cite{iccv21_Frozen}            &  CW &  32  & $28.9$ & $57.8$ &  $71.2$ &  $-$ \\
        SSB~\cite{iclr21_supportset}     &  H  &  - & $29.2$ & $61.6$ &  $-$ &  $-$ \\
        HiT~\cite{iccv21_HiT}            &  H &  -  & $29.6$ & $60.7$ &  $-$ &  $-$ \\
        CLIP4Clip~\cite{arxiv_clip4clip} (ViT-B/32) &  W &  64  & $40.7$ & $71.8$ &  $83.4$ &  $8.2$ \\ 
        \textbf{\EC} (ViT-B/32)     & W+V &  32 & $\textbf{42.3}$ & $\textbf{73.2}$ &  $\textbf{83.8}$ &  $\textbf{8.2}$ \\
        \textbf{\EC} (ViT-B/16)    & W+V &  32 & $\textbf{45.3}$ & $\textbf{75.7}$ &  $\textbf{86.2}$ &  $\bf 6.2$ \\
        \bottomrule
    \end{tabular}
\end{table}

\textbf{Comparison to the State-of-the-Art.}  In Table~\ref{tab:act_stoa}, we report the results of our method on ActivityNet Captions. These results indicate several interesting findings. First, we notice that the gap between CLIP-based methods (i.e., CLIP4Clip, \textsc{EclipSE}) and other previous approaches is significant ($>10\%$ in R@1 metric). This result justifies our motivation to build on the powerful CLIP model. Second, our results indicate that \EC outperforms CLIP4Clip by a substantial margin ($1.6\%$ in R@1), which suggests the usefulness of temporal audio cues. Third, we also note that unlike CLIP4Clip, which operates on $64$ frame inputs, \EC achieves higher accuracy while processing fewer frames (\ie $32$). Lastly, we show that using a stronger visual backbone (i.e., ViT-B/16 vs. ViT-B/32) leads to improved video retrieval performance.

\begin{wrapfigure}{h}{0.45\linewidth}
    \centering
\includegraphics[width=0.85\linewidth]{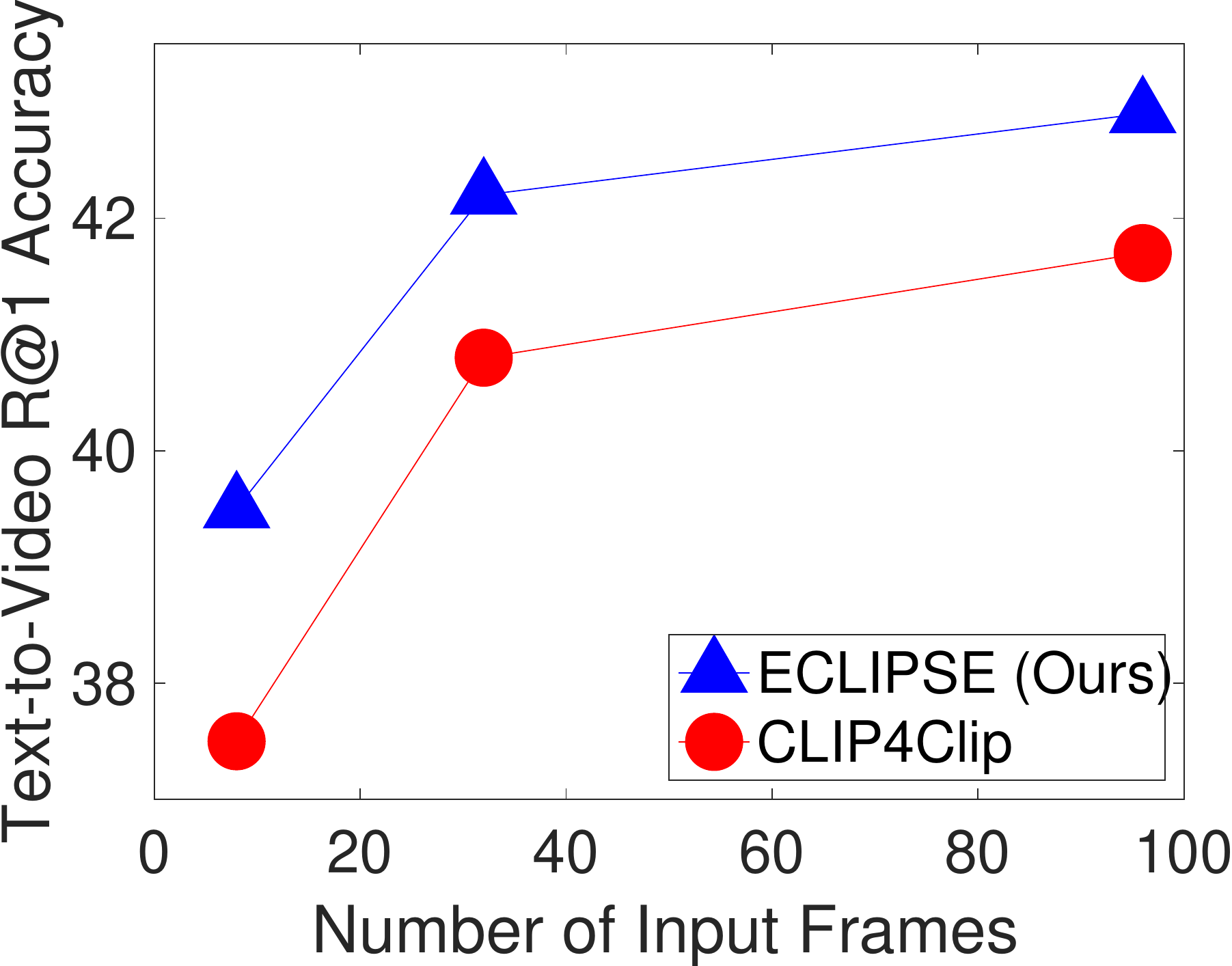}
    \caption{We compare \EC with CLIP4Clip with a varying number of frames. Our method outperforms CLIP4Clip while using the same number or even fewer frames.}
    \label{fig:num_frames2}
\end{wrapfigure}

\textbf{Accuracy vs. Number of Frames.} We next investigate the trade-off between video retrieval accuracy and the number of input frames. In Figure~\ref{fig:num_frames2}, we plot the long-range text-to-video retrieval accuracy (i.e., R@1) as a function of the number of input frames. 
Based on these results, we observe that \EC consistently outperforms CLIP4Clip in $8,32$ and $96$-frame regimes. Furthermore, we notice that \EC achieves higher accuracy than CLIP4Clip even when operating on a much smaller number of video frames (e.g., $32$ vs $96$). 

\begin{table}[b]
\footnotesize
    \caption{
    We compare the computational cost of a 32-frame \EC with a 96-frame CLIP4Clip~\cite{arxiv_clip4clip} on ActivityNet Captions. Both methods are built using a ViT-B/32 architecture. Despite using fewer frames, \EC outperforms CLIP4Clip. Additionally, our method uses $2.3\times$ less GPU memory, runs $2.92\times$ faster and is generally more efficient as indicated by the number of GFLOPs (i.e., $827$ vs $1251$).
    }
    \label{tab:comp_cost}
    \centering
    \begin{tabular}{l ccccc}
        \toprule
        {Method} & \multicolumn{1}{p{1.8cm}}{\centering Num. \\ Frames } & \multicolumn{1}{p{2.0cm}}{\centering Inference \\ GFLOPs $\downarrow$} & \multicolumn{1}{p{2.3cm}}{\centering GPU Mem. \\ (in MB) $\downarrow$} & \multicolumn{1}{p{1.8cm}}{\centering Samples \\  per Sec. $\uparrow$} & \multicolumn{1}{p{1.4cm}}{\centering T2V \\ R@1 $\uparrow$ }  \\  
        \midrule
        CLIP4Clip              & 96 & $1251$ & 24,802 &  $17.39$ &  $41.7$ \\
        \textbf{\EC}            & 32 & \bf 827    & \bf 10,637 &  \bf 50.93 &  \bf 42.3  \\
        \bottomrule
        
    \end{tabular}
\end{table}

\textbf{Computational Cost Analysis.} We note that compared to the video-only approaches,  our proposed \EC uses an additional audio modality. However, we would also like to emphasize that we use audio to improve the efficiency of the costly video-only approaches rather than merely improving the absolute video retrieval accuracy. In Table~\ref{tab:comp_cost}, we compare the computational cost of a 96-frame CLIP4Clip with our 32-frame \textsc{EclipSE}. Based on these results, we observe that \EC uses $2.3 \times$ less GPU memory, runs $2.92 \times$ faster, and achieves better accuracy ($42.3$ vs. $41.7$) than CLIP4Clip. This suggests that replacing the costly video modality with the audio makes our retrieval framework more efficient and also improves its accuracy.



\subsection{Results on Other Long-range Datasets}
\label{subsec:qvh_yc2}

Next, we validate our approach on four other long-range video datasets: QVHighlights~\cite{nips21_qvhighlights} (QVH), DiDeMo~\cite{didemo}, YouCook2~\cite{youcook2} (YC2), and Charades~\cite{eccv16_charades}. Since long-range video retrieval is a relatively unexplored subarea of research, we note that QVHighlights, YouCook2, and Charades are not formally used for the long-range video retrieval task. However, all three of these datasets contain (i) long videos and (ii) multiple annotated text descriptions of short-term segments within each long video. Thus, to re-purpose these datasets for long-range text-to-video retrieval, we follow the protocol of ActivityNet Captions~\cite{iccv17_activitynet}. Specifically, we concatenate the textual descriptions of all short-term segments in a given long video and treat it as a paragraph-to-video retrieval task similar to~\cite{iccv17_activitynet}. In our comparisons, we also include other recent video retrieval methods such as ClipBERT~\cite{cvpr21_clipbert}, Frozen in Time (FiT)~\cite{iccv21_Frozen}, and CLIP4Clip~\cite{arxiv_clip4clip}.

\begin{table}[t]
    \caption{
    Our results on QVHighlights~\cite{nips21_qvhighlights} (QVH), YouCook2~\cite{youcook2} (YC2), Charades~\cite{eccv16_charades} and DiDeMo~\cite{didemo} using the $R@1$ T2V metric. A 32-frame \EC with a ViT-B/32 backbone outperforms prior approaches while also being more efficient.
    }
    \label{tab:yc2_qvh_stoa}
    \centering
    \small
    \begin{tabular}{p{2.2cm} p{1.3cm} p{1.0cm} p{1.0cm} p{1.3cm} p{1.0cm} p{1.5cm} p{1.2cm}}
        \toprule
        Method &  Pretrain & Frames & QVH & DiDeMo & YC2 & Charades & GFLOPs \\
        \midrule
        ClipBERT~\cite{cvpr21_clipbert}      & C+G & 32 & $43.2$ & $20.4$ &  $29.8$ & $6.7$ & -   \\
        FiT~\cite{iccv21_Frozen}  & CW & 32 & $55.0$ & $35.8$ & $32.2$ &  $11.9$ & 1426 \\ 
        CLIP4Clip~\cite{arxiv_clip4clip} & W & 96 & $70.2$ & $42.5$ & $37.6$ &  $13.9$   & 1251\\
        \textbf{\EC}                                  & W+V & 32 & $\textbf{70.8}$ & $\textbf{44.2}$ & $\textbf{38.5}$ &  $\textbf{15.7}$ & \bf 827 \\
        \bottomrule
    \end{tabular}
\vspace{\tabmargin}
\end{table}


In Table~\ref{tab:yc2_qvh_stoa}, we show that a 32-frame \EC with a ViT-B/32 backbone outperforms prior methods on all four datasets. Additionally, we point out that our method is more efficient than both FiT~\cite{iccv21_Frozen} and CLIP4Clip~\cite{arxiv_clip4clip} (827 vs. 1251 vs. 1426 in GFLOPs).

\textbf{Potential Overlap between Audio and Video Datasets.} Next, we want to verify that the videos used to pretrain our audio encoders were not present in the test sets of the video retrieval benchmarks.  Upon our investigation, we found that the overlap between VGGSound, which was used to pretrain our best audio model, and ActivityNet Captions was small, i.e., $42$ out of $4,926$ videos ($0.8\%$). Furthermore, there were no overlaps between the VGGSound and all of the other datasets. To validate that our original conclusions on ActivityNet still hold, we conducted additional experiments on the deduplicated ActivityNet dataset where the overlapping test videos were removed. We used the same CLIP4Clip and \textsc{EclipSE} methods as in Table~\ref{tab:act_stoa}. We report that \textsc{EclipSE} achieved $42.3\%$ T2V R@1 accuracy while CLIP4Clip obtained $40.8\%$. Both of these results are almost identical to the results in Table~\ref{tab:act_stoa} (i.e., $42.3\%$ and $40.7\%$ respectively). 

\subsection{Ablation Studies}

Next, we investigate how different design choices of our model affect the long-range video retrieval accuracy on the ActivityNet Captions dataset~\cite{iccv17_activitynet}.

\textbf{Audiovisual Block Design.} First, we validate the effectiveness of our audiovisual attention block by comparing it to (i) a joint audiovisual attention that processes concatenated video and audio tokens, (ii) the variant of our model that only uses audio-to-video (A2V) attention (\eqnref{a2v_att}) and (iii) our final model that uses both audio-to-video (A2V) and video-to-audio (V2A) attentions (\eqnref{a2v_att} and \eqnref{v2a_att}). For efficiency, all models are trained using 8-frame inputs.

\begin{figure}[t!]
    \centering
    \includegraphics[width=1\linewidth]{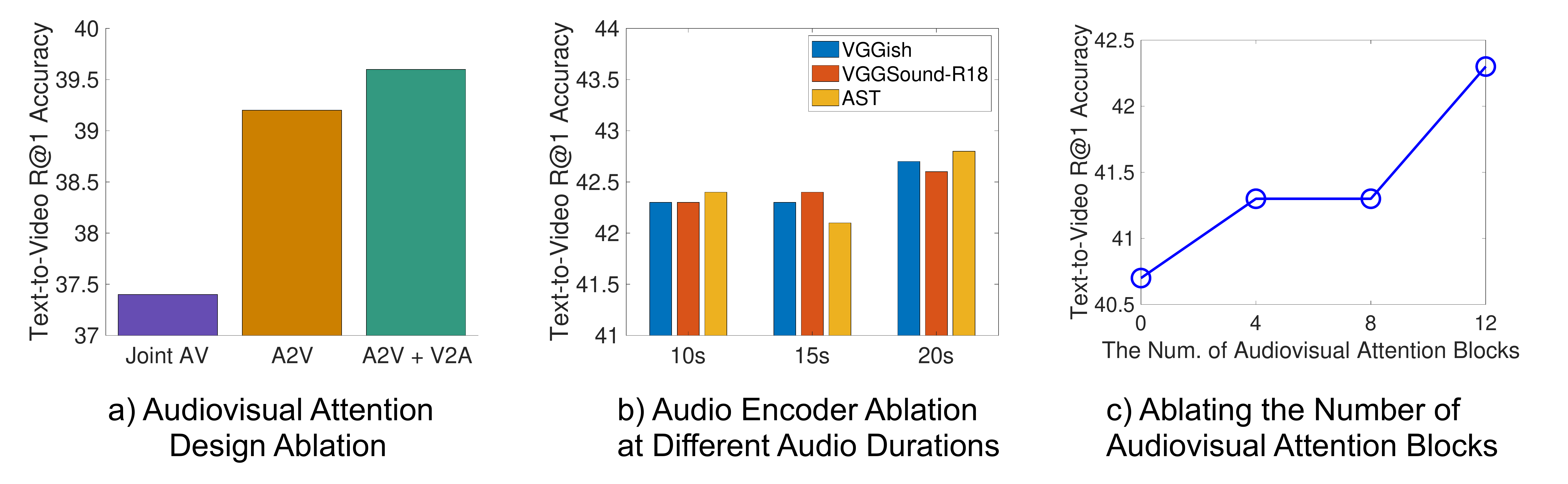}
	\vspace{\figmargin}
    \caption{\textbf{(a)} In the left subfigure, we study different audiovisual block design. Joint AV refers to standard self-attention applied to concatenated audio and video tokens. A2V refers to a single cross-modal audio-to-video attention block (\eqnref{a2v_att}). Lastly, A2V+V2A depicts our dual-pathway attention block design (\eqnref{a2v_att} and \eqnref{v2a_att}). Based on these results, we observe that dual-pathway attention achieves the best performance. For efficiency, we use $8$ frame inputs for these experiments. \textbf{(b)} In the middle subfigure, we also investigate different audio encoders applied to different duration audio segments. These results indicate that (i) longer audio typically improves the performance, (ii) \EC is robust to different audio encoders. \textbf{(c)} In the right subfigure, we study video retrieval accuracy as a function of the number of audiovisual attention blocks. Based on these results, we observe that injecting our proposed audiovisual attention block into every layer of our 12-layer \EC model leads to the best performance.
    }
    \label{fig:ablation}
    \vspace{\figmargin}
\end{figure}
\begin{figure}[t!]
    \centering
	\includegraphics[width=0.95\linewidth]{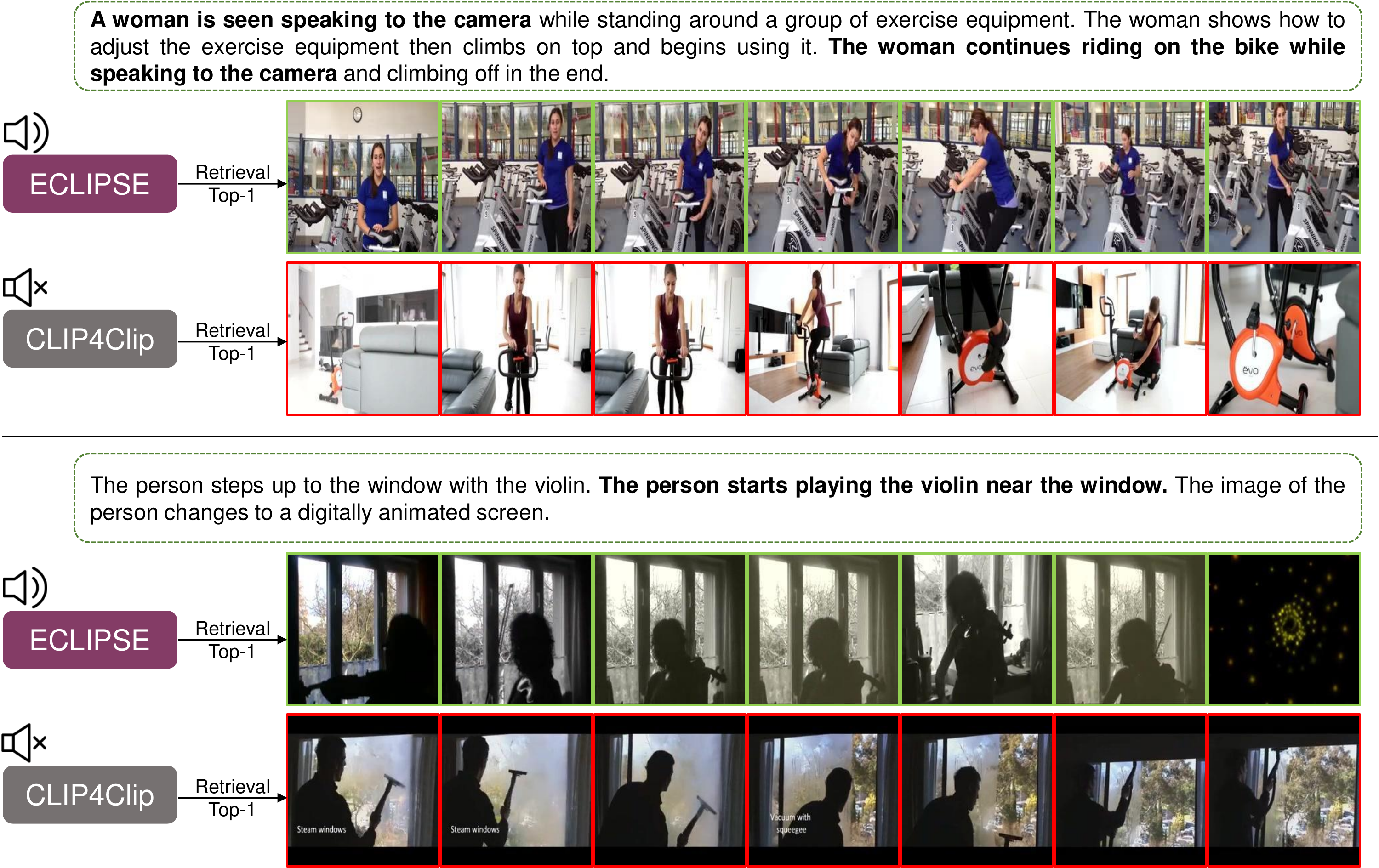}
	\vspace{\figmargin}
    \caption{
    Here, we illustrate our qualitative retrieval results on ActivityNet Captions~\cite{iccv17_activitynet}. We compare our audiovisual \EC model with a video-only CLIP4Clip~\cite{arxiv_clip4clip}. 
    %
    For a given a textual query (depicted in a green block), we visualize each method's top-1 retrieved video. Our results indicate that the video-only CLIP4Clip struggles with retrieval when textual queries include  audio event descriptions, e.g., ``a woman speaking to the camera", ``a person playing the violin," etc. (see bolded text). In these cases, CLIP4Clip fails to retrieve the correct video instances, whereas \EC effectively leverages audiovisual cues for a successful retrieval.
    }
	\label{fig:vis}
	\vspace{\figmargin}
\end{figure}

From the results in \figref{ablation}a, we observe that using a bi-directional audiovisual attention (i.e., both audio-to-video (A2V) and video-to-audio (V2A)) leads to the best $R@1$ text-to-video retrieval accuracy on ActivityNet. 

\textbf{Different Audio Encoders.}
Next, we study how different audio encoders affect the video retrieval performance of our model. Specifically, we experiment with CNN-based audio encoders such as VGGish~\cite{icassp17_vggish} and VGGSound~\cite{icassp20_vggsound}, and also a transformer-based audio encoder AST~\cite{TASLP_AST}. 
Our results in \figref{ablation}b suggest that our framework is robust to the choice of an audio encoder, as all three audio encoders produce a similar performance.

\textbf{Audio Duration.} Additionally, we investigate how audio duration affects the accuracy of a long-range video retrieval task. 
In \figref{ablation}b, we experiment with audio spectrograms of 10, 20, and 30 seconds.
Our results indicate that longer audio duration leads to better performance. However, the performance gain is relatively small (i.e., $0.5\%$ in R@1), suggesting that $10s$ audio spectrograms are typically sufficient to capture relevant audio cues.

\textbf{The Number of Audiovisual Attention Blocks.} 
In \figref{ablation}c, we also study the video retrieval performance (using R@1) as a function of the number of audiovisual attention blocks in our 12-layer \EC model. Using $k$ audiovisual attention blocks implies that these $k$ audiovisual blocks are injected into the first $k$ layers of the network while the remaining $12 - k$ layers only consider visual information. 
Our results indicate that video retrieval performance decreases when we use fewer audiovisual attention blocks.
In other words,  our method achieves the best video retrieval accuracy when the audiovisual attention block is inserted into every layer of our \EC architecture.
%



\textbf{The Importance of CLIP Pretraining.}  To highlight the importance of CLIP pretraining, we compare CLIP pretraining with the ImageNet-21k pretraining. We use the ViT-B/32 architecture for these experiments. We report that compared to the ImageNet-21k pretraining, CLIP pretraining leads to $27.1\%$, $34.6\%$, $24.2\%$, $42.5\%$ better T2V R@1 retrieval accuracy on ActivityNet, DiDeMo, YouCook2, and QVHighlights respectively. These results suggest that CLIP pretraining is essential for good downstream video retrieval performance. 

\textbf{Single Modality Baselines.} We also report the results of (i) 180-second audio-only, (ii) 64-frame video-only, and (iii) our 32-frame audiovisual methods. On ActivityNet Captions, the three approaches achieve $2.7\%$, $40.7\%$, and $42.3\%$ R\@1 T2V retrieval accuracy respectively. These results indicate that jointly modeling audio and video achieves the best accuracy. We also note that while audio alone obtains poor accuracy, audio effectively complements video in our audiovisual approach. We observe similar trends on all other datasets too. 



%
%

\begin{figure}[t!]
    \centering
	\includegraphics[width=0.83\linewidth]{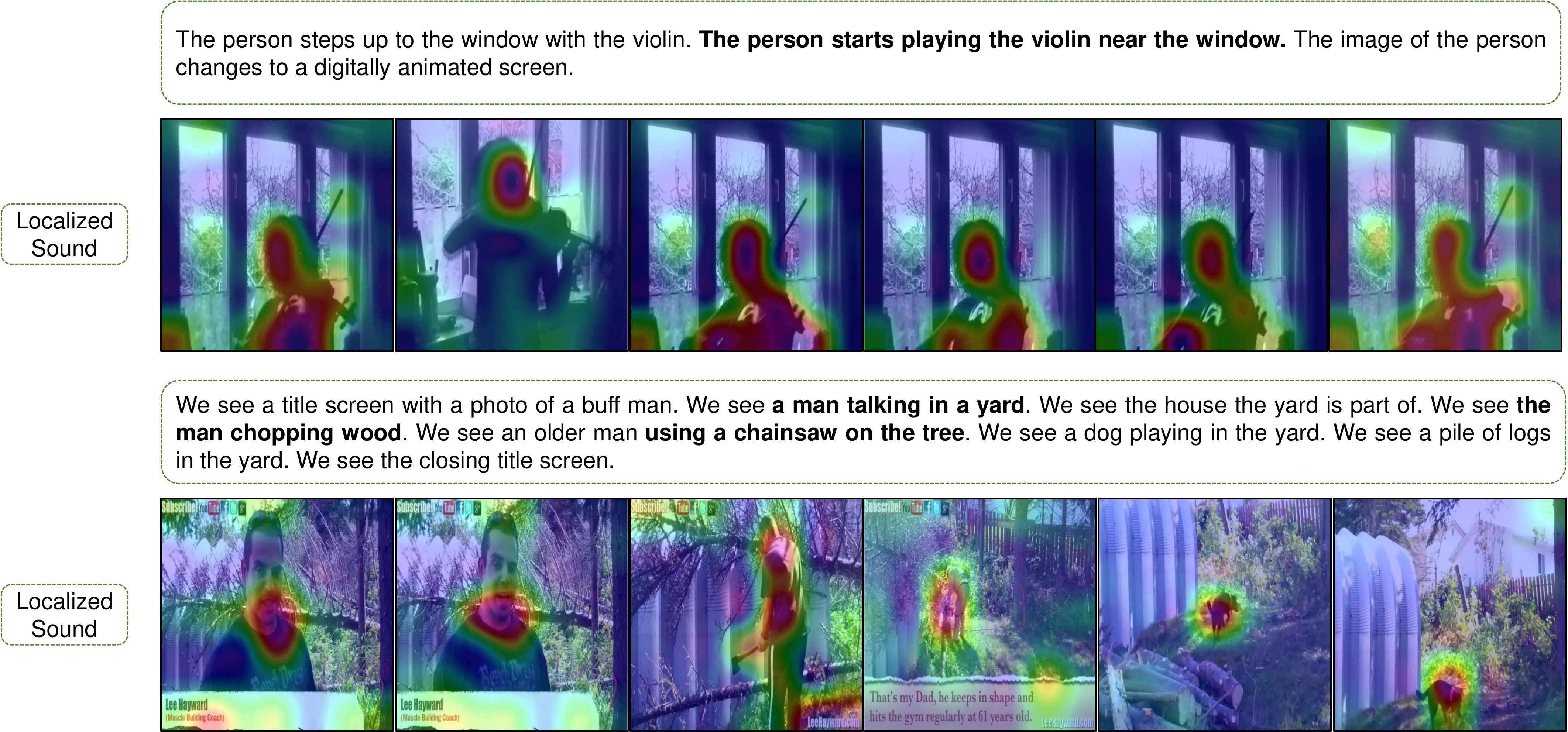}
	\vspace{\figmargin}
    \caption{
    Here, we illustrate qualitative sound localization results of our method. Note that our \EC is not explicitly trained for the sound localization task.
    In other words, \EC learns implicit associations between objects and sounds while being optimized with respect to the video retrieval task.
    }
	\label{fig:vis_sound}
	\vspace{\figmargin}
\end{figure}

\subsection{Qualitative Results}

\textbf{Video Retrieval Results.} In \figref{vis}, we also illustrate some of the qualitative video retrieval results on ActivityNet Captions~\cite{iccv17_activitynet}.
Specifically, for a given textual query (illustrated in the green blocks in~\figref{vis}), we visualize the top-1 retrieved video by our audiovisual \EC and the video-only CLIP4Clip baseline.
Based on these results, we observe that the video-only CLIP4Clip method struggles to retrieve videos for textual queries that include audio-based event descriptions. For instance, in the first example of \figref{vis}, the textual query mentions an audio-based event of ``a woman speaking to the camera" (see bolded text). Furthermore, the textual query in the second example also involves a sound event of ``a person playing the violin." 

Since CLIP4Clip, does not have any audiovisual modeling capabilities, it fails to retrieve the correct video in these cases. In contrast, \EC retrieves the correct videos in all three illustrated cases, thus, highlighting the importance of incorporating video and audio cues for effective long-range video retrieval.

\textbf{Sound Localization Results.} In \figref{vis_sound}, we also demonstrate qualitative sound localization results of our method. 
By computing the similarity between audio features and visual patches, we can obtain saliency maps that are indicative of sound sources in the video.
Note that our method does not require any additional sound localization training objective.
In other words, \EC successfully learns associations between sound sources and objects (\eg a woman talking, a man playing the violin, a man using a chainsaw) as a byproduct of being trained for the video retrieval task.


\section{Conclusions}
In this paper, we present a novel audiovisual framework,~\textsc{EclipSE}, for long-range video retrieval.
By replacing costly and redundant parts of the video, with compact audio cues, \EC efficiently processes long-range videos while also obtaining better performance than standard video-only methods. 
Our audiovisual framework is (i) flexible, (ii) fast, (iii) memory-efficient, and (iv) it achieves state-of-the-art results on five diverse long-range video benchmarks.
In the future, we plan to extend our method to other multimodal video understanding tasks such as video question answering and video captioning. 

\clearpage
\newcount\cvprrulercount
\appendix

\section{Appendix}
Our appendix consists of:
\begin{enumerate}
    \item Implementation Details.
    \item Additional Quantitative Results.
    \item Additional Qualitative Results. 
    \item A Supplementary Video. 
\end{enumerate}

\sloppy
\section{Additional Implementation Details}
\textbf{Experimental Setting.} In all experiments, the visual frames are extracted at $3$ fps. 
We adopt pretrained CLIP~\cite{icml21_clip} on both text and visual encoder, which is based on the ViT-B/32 visual backbone.
We initialize the weights of our proposed audiovisual block using the corresponding spatial attention weights of CLIP.
To gradually incorporate audio information into visual features, we attach a learnable fully connected layer to each audiovisual attention block and initially set it to zero.
For visual representations, we first "patchify" $224 \times 224$ video frames into $32 \times 32$ patches as is done in~\cite{icml21_clip}.
Each video frame is then tokenized into $49$ patches and a learnable $768$-dimensional  \textit{CLS} token.
At the end, the frame-level \textit{CLS} tokens are averaged to obtain a video-level feature embedding that is used to optimize our model as described in~\secref{loss_func}.
For audio encoder, we use ResNet-18~\cite{resnet} pre-trained on VGGSound~\cite{icassp20_vggsound}.
We sample 10-second audio clips in the neighborhood around the sampled video frame and process the raw audio into spectrogram as is done in~\cite{icassp20_vggsound}.
Lastly, for textual features, we adopt CLIP tokenizer for all text inputs.
Specifically, the textual encoder processes all textual tokens and a special $768$-dimensional  \textit{CLS} token as its inputs.
Afterward, we only consider the \textit{CLS} textual token to match a given video with the corresponding textual description.

\noindent\textbf{Training Details.}
We implement \EC using Pytorch~\cite{pytorch} and conduct the training on four NVIDIA A6000 GPUs. 
For fair comparison with the baseline methods, we set the batch size to 64.
We train our model with Adam optimizer~\cite{adam} and set the learning rate to $1e-7$ for text encoder and spatial attention in \eqnref{spatial_att} of main paper with weight decay $0.2$.
For our audiovisual attention blocks, A2V and V2A (see \eqnref{a2v_att} and \eqnref{v2a_att} in our main draft).
The maximum text input is set to $64$ tokens for Charades, DiDeMo, and QVHighlight. 
We set $128$ for ActivityNet Captions and YouCook2 due to longer paragraph.

\section{Additional Quantitative Results}

\begin{table}[t]\centering
\caption{We investigate how different video frame sampling strategies affect the performance of a \textbf{video-only} CLIP4Clip~\cite{arxiv_clip4clip} baseline on ActivityNet Captions~\cite{iccv17_activitynet}. The results are reported in text-to-video $R@1$  metrics. We observe that for a smaller number of frames (e.g., 32-64) random sampling yields slightly better performance than the uniform sapling. Conversely, for a larger number of frames (e.g., 96-128) uniform sampling leads to better accuracy.}\label{tab:sample}
\begin{tabular}{lrrrrr}\toprule
\diagbox[width=12em]{Method}{ Num. Frames } &32 &64 &96 &128 \\\cmidrule{1-5}
Uniform &40.4 &40.7 & \bf 41.7 &40.9 \\\cmidrule{1-5}
Random Sample &41.0 &41.2 &40.9 &40 \\
\bottomrule
\end{tabular}
\vspace{\tabmargin}
\end{table}

\noindent\textbf{Ablating Different Frame Sampling Strategies.} In \tabref{sample}, we investigate different video frame sampling strategies on ActivityNet Captions using R@1 evaluation metric.
Specifically, we experiment with uniform and random frame sampling using a CLIP4Clip baseline~\cite{arxiv_clip4clip}.
For uniform sampling, we sample the frames uniformly throughout the entire input video.
For random sampling, we divide the video into a fixed number of segments, and randomly sample one frame within each segment.
Based on the results in \tabref{sample}, we note that random sampling improves performance for a smaller number of video frames (\eg 32-64).
Conversely, when using a larger number of frames (\eg 96-128) the uniform sampling strategy leads to slightly better accuracy.
For simplicity, we use the standard uniform sampling strategy for all of our experiments.

\noindent\textbf{Comparison with Frozen in Time (FiT)~\cite{iccv21_Frozen}.}
In addition to the comparisons with FiT in \tabref{act_stoa} and \tabref{yc2_qvh_stoa} of the main draft, here, we include more detailed comparisons on ActivityNet (\textbf{Act}), DiDeMo (\textbf{DD}), YouCook2 (\textbf{YC2}) and QVHighlights (\textbf{QVH}). Overall, the results below indicate that compared to FiT, \textsc{EclipSE} achieves better accuracy on all datasets and it also has fewer GFLOPs for the same number of input frames.

\begin{table}[h]
    \caption{
    Frozen in Time (FiT)~\cite{iccv21_Frozen} and our results on on ActivityNet (\textbf{Act}), DiDeMo (\textbf{DD}), YouCook2 (\textbf{YC2}) and QVHighlights (\textbf{QVH}) using the $R@1$ metric. \EC outperforms FiT while also being more efficient.
    }
    \centering
    \small
    \begin{tabular}{p{2.4cm} p{1.4cm} p{1.4cm} p{1.4cm} p{1.4cm} p{1.4cm} p{1.4cm}}
        \toprule
        Method  & Frames & Act & DD& YC2 & QVH & GFLOPs \\
        \midrule
        FiT~\cite{iccv21_Frozen}   & 8  & $24.8$ & $34.6$ & $21.2$ &  $41.2$ & $357$ \\ 
        \textbf{\textsc{EclipSE}}           & 8 &  $\bf 39.6$ & $\bf 40.4$   & $\bf 28.8$ & $\bf 52.1$ & \bf 313\\ \hline
        FiT~\cite{iccv21_Frozen}   & 32 & $28.9$ & $35.8$    & $32.2$ &  $55.0$ & $1426$ \\ 
        \textbf{\textsc{EclipSE}}           & 32 & \bf 42.3 & \bf 44.2    & \bf 38.5 &  \bf 70.8 & \bf 827 \\ 
        \bottomrule
    \end{tabular}
\end{table}

\section{Additional Qualitative Results}

\begin{figure}[t!]
    \centering
	\includegraphics[width=0.8\linewidth]{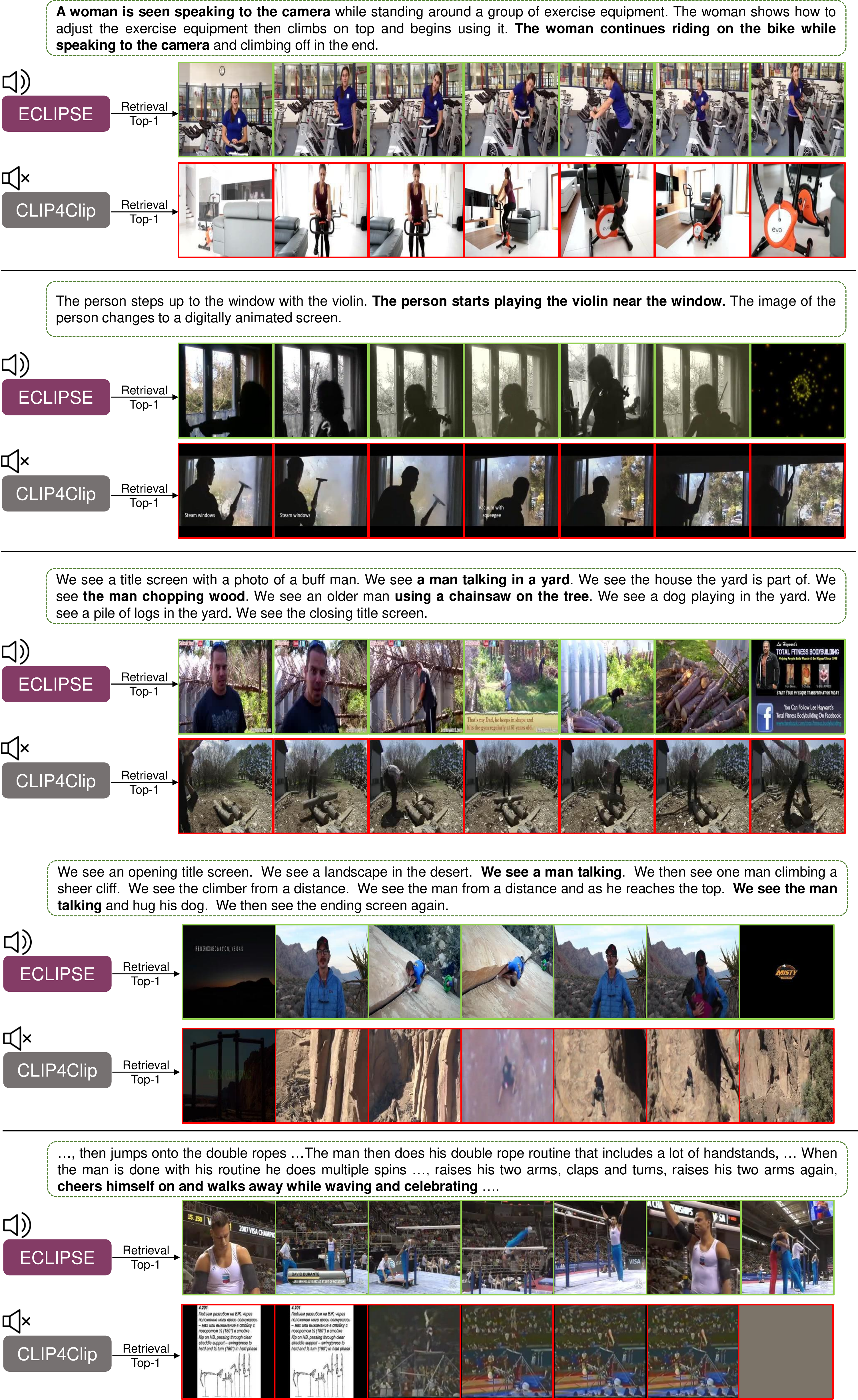}
	\vspace{\figmargin}
    \caption{
    Here, we illustrate our qualitative long-range retrieval results on ActivityNet Captions~\cite{iccv17_activitynet}. We compare our audiovisual \EC model with a video-only CLIP4Clip~\cite{arxiv_clip4clip}.
    For a given a textual query (depicted in a green block), we visualize each method's top-1 retrieved video. Our results indicate that the video-only CLIP4Clip struggles with retrieval when textual queries include  audio event descriptions, e.g., ``a man talking", ``a person cheering," etc. (see bolded text). In these cases, CLIP4Clip fails to retrieve the correct video instances, whereas \EC effectively leverages audiovisual cues for successful long video retrieval.
    }
	\label{fig:vis_supp_in_app}
	\vspace{\figmargin}
\end{figure}

\noindent\textbf{Video Retrieval Results.} In \figref{vis_supp_in_app}, we provide additional qualitative results of our long-range video retrieval framework on ActivityNet Captions~\cite{iccv17_activitynet}.
%
%
In all of these examples, we notice that CLIP4Clip baseline fails to capture relevant audio-based events (\eg people cheering). In comparison, our \EC model successfully retrieves videos that contain complex audiovisual events, thus, highlighting the importance of audiovisual modeling for long-range video retrieval. 

\begin{figure}[t!]
    \centering
	\includegraphics[width=0.83\linewidth]{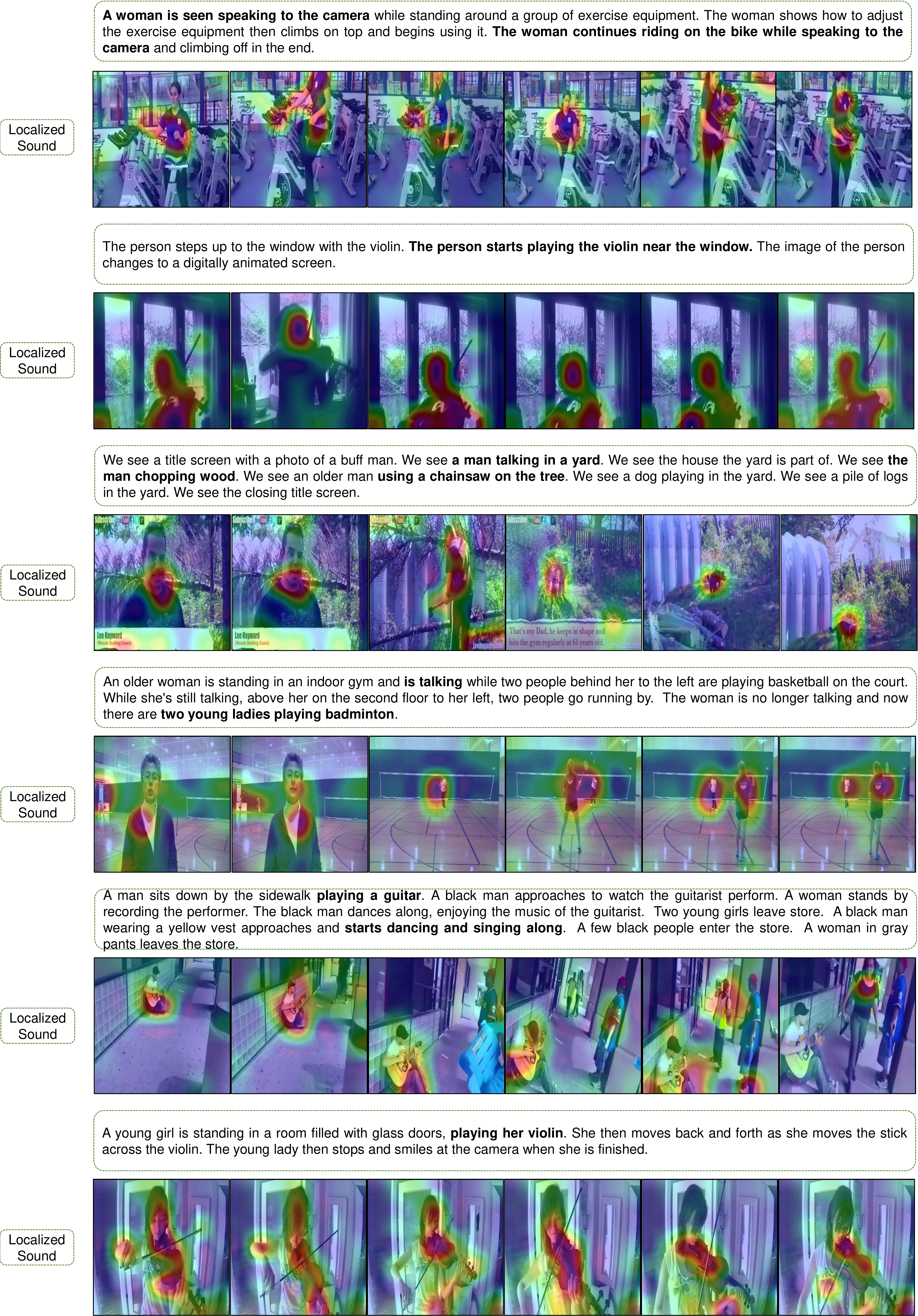}
	\vspace{\figmargin}
    \caption{
    Here, we illustrate qualitative sound localization results of our method. Note that our \EC is not explicitly trained for the sound localization task.
    In other words, \EC learns implicit associations between objects and sounds while being optimized with respect to the video retrieval task.
    }
	\label{fig:vis_sound_supp}
	\vspace{\figmargin}
\end{figure}

\noindent\textbf{Sound Localization Results.} In \figref{vis_sound_supp}, we also demonstrate qualitative sound localization results of our method. 
Specifically, by computing the similarity between audio features and visual patches, we can obtain saliency maps that are indicative of sound sources in the video.
Furthermore, we would like to emphasize that our \EC model does not require any additional sound localization training objective.
In other words, \EC successfully learns associations between sound sources and objects (\eg a woman talking, a man playing the violin, a man using a chainsaw) as a byproduct of being trained for the video retrieval task.

\section{Supplementary Video}

Lastly, our appendix also includes a video (see our project page) illustrating our qualitative results in the video format. Specifically, we include the results of our \EC model on several challenging video retrieval cases. For comparison, we also include the results of a CLIP4Clip baseline. Additionally, in these video results, we demonstrate that \EC also learns to localize sounds in the video even though it was not explicitly trained to do so. Overall, our video results indicate that compared to CLIP4Clip, \EC is more robust when retrieving long videos particularly in cases that involve complex audiovisual events.

\clearpage
%
%
\bibliographystyle{unsrt}
\bibliography{egbib}
\end{document}